\pdfoutput=1

\documentclass[11pt]{article}

\usepackage{acl}

\usepackage{times}
\usepackage{latexsym}

\usepackage[T1]{fontenc}

\usepackage[utf8]{inputenc}

\usepackage{microtype}

\usepackage{inconsolata}

\usepackage{graphicx}

\usepackage{amsmath}
\usepackage{mathtools}

\DeclareMathOperator*{\argmin}{argmin}

\usepackage{pifont}
\usepackage{graphicx}
\usepackage{enumitem}
\usepackage{pgfplots}
\usepackage{hyperref}
\usepgfplotslibrary{fillbetween}
\pgfplotsset{width=10cm,compat=1.9}

\usepackage{array}
\usepackage{booktabs}
\usepackage{multirow}
\usepackage{multicol}
\usepackage{tabularx}
\newcolumntype{C}{>{\centering\arraybackslash}X}

\usepackage{bm}
\usepackage{xspace}
\usepackage{xcolor}
\definecolor{softred}{RGB}{250,100,100}
\definecolor{softgreen}{RGB}{56,118,29}
\definecolor{softblue}{RGB}{100,150,200}
\newcommand{\textred}[1]{{\color{softred}#1}}
\newcommand{\textblue}[1]{{\color{softblue}#1}}
\newcommand{\textgreen}[1]{{\color{softgreen}#1}}
\newcommand{\ourmethod}{\textsc{Fused}\xspace}
\newcommand{\ourmetric}{\textsc{DM}\xspace}

\title{Improving Demonstration Diversity by Human-Free Fusing \\ for Text-to-SQL}

\author{
    Dingzirui Wang, Longxu Dou, Xuanliang Zhang, Qingfu Zhu, Wanxiang Che \\
    Harbin Institute of Technology \\
    \{dzrwang, lxdou, xuanliangzhang, qfzhu, car\}@ir.hit.edu.cn
}

\begin{document}
    \maketitle

    \begin{abstract}
        In-context learning with large language models (LLMs) is the current mainstream method for text-to-SQL. 
        Previous studies have explored selecting relevant demonstrations from a human-labeled demonstration pool, but these methods lack diversity and incur high labeling costs.
        In this work, we address measuring and enhancing the diversity of the text-to-SQL demonstration pool. 
        First, we introduce a diversity metric and present that the diversity of the existing labeling data can be further enhanced.
        Motivated by these findings, we propose \ourmethod that iteratively fuses demonstrations to create a diverse demonstration pool based on human labeling or even from scratch with LLMs, reducing labeling costs.
        \ourmethod achieves an average improvement of $3.2\%$  based on existing labeling and $5.0\%$ from scratch on several mainstream datasets, demonstrating its effectiveness.\footnote{Our data and code are released in \href{https://github.com/zirui-HIT/Fused.git}{https://github.com/zirui-HIT/Fused}.}
    \end{abstract}

    \section{Introduction}
        \begin{figure}[t]
    \centering
    \includegraphics[width=\linewidth]{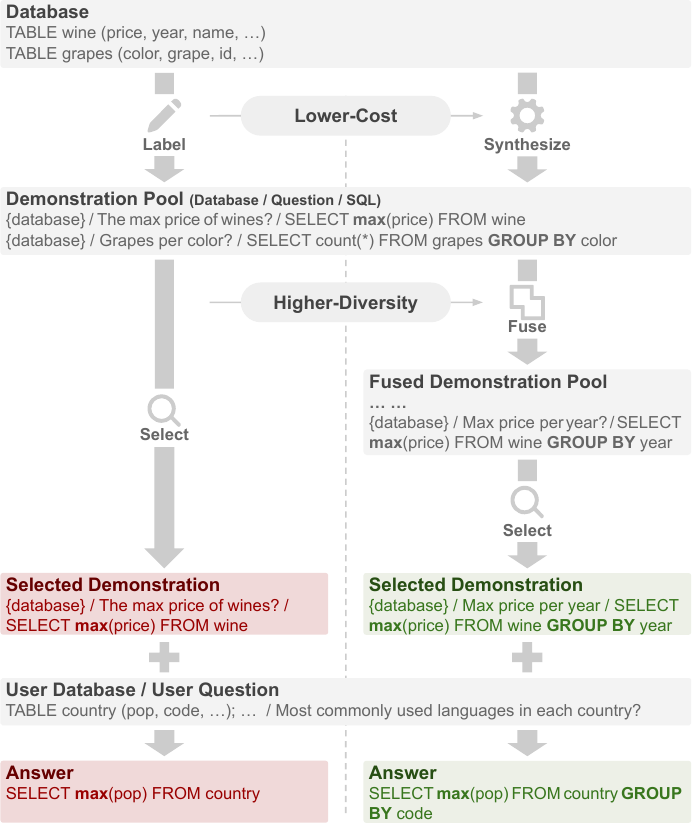}
    \caption{
        The comparison between the baseline (left) and \ourmethod (right) of obtaining the demonstration pool for text-to-SQL.
        \ourmethod can synthesize the demonstration pool from scratch or enhance the diversity of the existing labeling without additional human involvement.
    }
    \vspace{-1em}
    \label{fig:motivation}
\end{figure}

Text-to-SQL is a critical task that reduces the overhead of accessing databases by automatically generating SQL queries in response to user questions \cite{qin-etal-2020-SQLSurvey}. 
Recently, in-context learning based on large language models (LLMs) has become the predominant method for this task, significantly improving performance while minimizing the need for fine-tuning \cite{chen2024opensql,qu2024tasql,talaei2024chess}. 
For the in-context learning paradigm, besides the user question and the database, the LLM is also provided with several demonstrations, guiding the model to generate the corresponding SQL queries accurately.

Currently, numerous works \cite{su2023selective,ren2024purple,pourreza2024sqlencoder} explore how to select question-relevant demonstrations from a human-labeled demonstration pool. 
However, relying entirely on human labeling limits the performance of text-to-SQL based on in-context learning due to two main issues:
\textit{(i) Low Diversity}: Human-labeled data could lack diversity since the data labeled by the same annotator could be somewhat similar \cite{ramalingam2021more,guo2024relabel};
\textit{(ii) High Cost}: Human labeling requires significant labor overhead.
To address these issues, thereby improving text-to-SQL performance, we discuss:
\textbf{\textit{(i) Theoretical metric for measuring the diversity of the demonstration pool}} (\S\ref{sec:discussion});
\textbf{\textit{(ii) Practical method that builds a diverse demonstration pool with existing labeling or even from scratch}} (\S\ref{sec:method}).

First, we analyze that the diversity of the existing labeling can be further enhanced. 
We begin by discussing the necessity of demonstration pool diversity and present a diversity metric called \textbf{Diversity Measurement (\ourmetric)}.
Using the metric, we prove that the existing labeling diversity can be further enhanced by showing that there exist demonstration pools with significantly higher \ourmetric.

Based on this analysis, we present our method called \textbf{FUS}ing it\textbf{E}ratively for \textbf{D}emonstrations (\ourmethod), which iteratively synthesizes the demonstrations using LLMs with existing labeling or from scratch, as shown in Figure~\ref{fig:motivation}. 
To tackle the \textit{Low Diversity}, \ourmethod fuses demonstrations from previous iterations, ensuring that the new demonstrations are distinct from the previous, thus enhancing diversity.
To address the \textit{High Cost} of labeling, our method employs LLMs to generate demonstrations, thereby reducing the need for human labeling. 

To validate the effectiveness of our method, we apply \ourmethod to several mainstream text-to-SQL datasets, including Spider~\cite{yu-etal-2018-spider} and KaggleDBQA~\cite{lee-etal-2021-kaggledbqa}. 
We synthesize demonstrations and compare performance with existing labeling and from scratch, where \ourmethod achieves an average performance improvement of $3.2\%$ and $5.0\%$, respectively, confirming its effectiveness. 
Further analysis shows that \ourmethod significantly enhances \ourmetric of the existing labeling, demonstrating its capability to enhance the diversity of the existing demonstration pool.

Our contributions are as follows:
\begin{itemize}[nolistsep,leftmargin=*]
    \item We present \textbf{\ourmetric}, a metric to measure the diversity of a given demonstration pool for text-to-SQL, revealing that the diversity of the existing human-labeling data can be further enhanced.
    \item We propose \textbf{\ourmethod}, a method to build a high-diversity demonstration pool iteratively through human-free synthesis based on existing labeling data or even from scratch.
    \item We validate \ourmethod on multiple mainstream text-to-SQL datasets, achieving performance improvements of \textbf{3.2\%} with existing labeling and \textbf{5.0\%} from scratch, demonstrating its effectiveness.
\end{itemize}


    \section{Analysis}
        \label{sec:discussion}
        \begin{figure}
    \centering
    \includegraphics[width=0.85\linewidth]{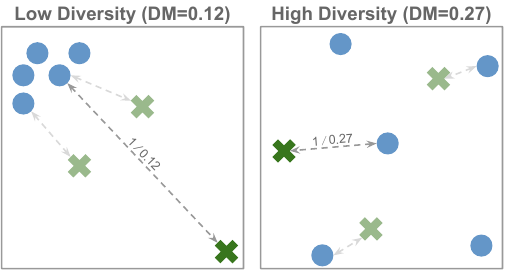}
    \caption{
        Two demonstration pools with different \ourmetric. 
        \textblue{\bm{$\bullet$}} represents the encoded demonstration, and \textgreen{\ding{54}} represents the encoded user questions, in which the darkest denotes the user question with the least similarity to the most similar demonstration. 
        The Euclidean distance between the user question and the most similar demonstration is indicated next to each line.
    }
    \vspace{-1em}
    \label{fig:demonstration_diveristy}
\end{figure}

In this section, we present that \textbf{the diversity of the existing labeled demonstration pool can be further enhanced}.
First, we discuss the necessity of high diversity for a demonstration pool.
Then, we introduce a metric to quantify the demonstration pool diversity.
Based on this metric, we discuss that the diversity of the existing labeling data can be further enhanced.
We compare the metric present with the other existing metric in Appendix~\ref{app:other_metric_comparison}.

\paragraph{Necessity of the Diversity}
    Regarding in-context learning, LLMs imitate the demonstration provided to generate the answer \cite{gpt3}.
    Therefore, given a user question, previous works select the most similar demonstrations from a demonstration pool to guide the LLMs in generating answers \cite{luo2024incontext-survey}.
    However, since user questions are unpredictable, the demonstration pool should be as diverse as possible to cover various user questions.
    The higher the diversity, the higher the similarity between any user questions and the demonstrations, thereby better guiding the answer generation; the lower the diversity, the more and more user questions are less similar to the demonstrations, decreasing the model performance.

\begin{figure*}[ht]
    \centering
    \includegraphics[width=\textwidth]{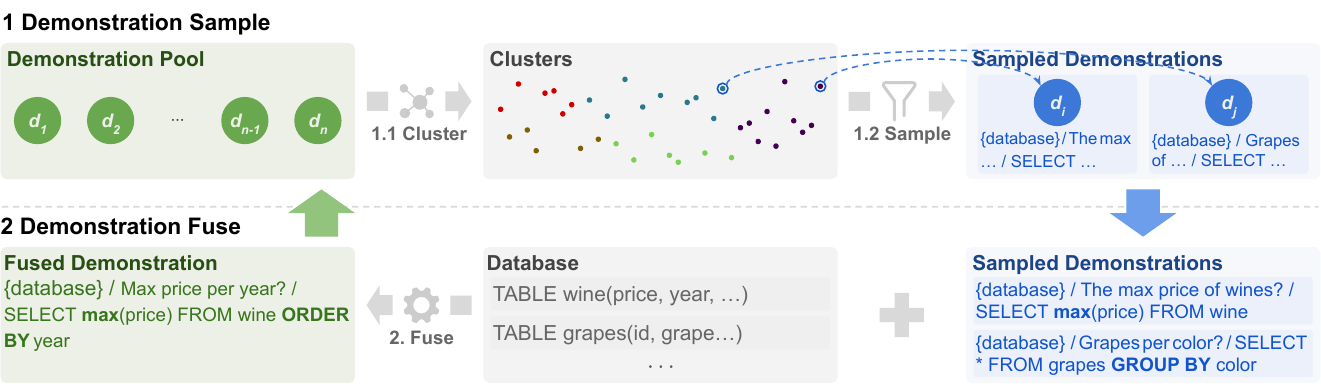}
    \caption{
        The pipeline of \ourmethod, which consists of two steps:
        \textbf{\textit{(i)} Demonstration Sample}: Sample demonstrations to be fused from the demonstration pool;
        \textbf{\textit{(ii)} Demonstration Fuse}: Fuse the sampled demonstrations with the randomly sampled database.
        The representation of \texttt{\{database\}} is discussed in Appendix~\ref{app:prompts}.
    }
    \vspace{-1em}
    \label{fig:pipeline}
\end{figure*}

\paragraph{Diversity Measurement}
    Based on the preceding discussion, we employ the user question with the lowest similarity to the demonstration pool to measure the diversity of the demonstration pool. 
    Formally, let $D = \{d_i\}$ represent the demonstration pool, $U = \{u\}$ denote the user questions, and $\texttt{sim}(u, d)$ as the similarity between $u$ and $d$, calculated as the reciprocal of the Euclidean distance between their encoded vectors in this paper. 
    We utilize Equation~\ref{equ:diversity_measurement} to measure the diversity of the demonstration pool $D$, which is called \textbf{Diversity Measurement (\ourmetric)}. 
    This metric corresponds to the similarity of the user question with the least similarity to the most similar demonstration in the demonstration pool, compared with any other user question. 
    An illustration of the \ourmetric definition is shown in Figure~\ref{fig:demonstration_diveristy}. 
    The detailed definitions of $U$, $\texttt{sim}$, and the calculation process of \ourmetric are discussed in Appendix~\ref{app:measurement}.

    \begin{equation}
        \ourmetric = \min_{u \in U} \max_{d_i \in D} \texttt{sim}(u, d_i)
        \label{equ:diversity_measurement}
    \end{equation}

\paragraph{Diversity of the Existing Labeling Can be Further Enhanced}
    With the metric present above, we then measure the diversity of the existing text-to-SQL labeling demonstration pool. 
    The DM and performance of the existing labeling demonstration pool are depicted in Figure~\ref{fig:diversity_label} and Figure~\ref{fig:diversity_measurement}. 
    These figures reveal other demonstration pools where DM and performance are significantly higher than the existing labeling data. 
    Thus, although the existing labeling exhibits relatively high diversity, it can be further improved, thereby enhancing the performance. 
    Consequently, we next discuss the method for synthesizing demonstrations to enhance the diversity of the demonstration pool.

    \section{Method}
        \label{sec:method}
Our method focuses on how to synthesize new demonstrations given databases with LLMs.
Considering the poor diversity of directly generating demonstrations only relying on the sampling generation \cite{cegin2024effects}, we present to synthesize by fusing different demonstrations iteratively, as shown in Figure~\ref{fig:pipeline}.
In each iteration, we guide the model to generate demonstrations that are not similar to the previous iterations, thereby enhancing the diversity.
We theoretically prove that our method can enhance \ourmetric in Appendix~\ref{app:provment_of_improvement}.

A simplified explanation of our method is that:
we first cluster the demonstrations based on the SQL keywords (e.g., \texttt{WHERE}, \texttt{ORDER BY}).
Then, we sample and fuse demonstrations from each cluster.
The fused demonstration contains both \texttt{WHERE} and \texttt{ORDER BY} that are different from the sampled demonstrations, thereby enhancing the demonstration diversity.
In practice, we use the encoded user questions rather than SQL keywords for synthesis since the user question has more semantic information than the SQL \cite{qin-etal-2020-SQLSurvey}.

\subsection{Overview}
    The fusion process of \ourmethod starts with an initial demonstration pool, which can be human-labeled or synthesized from scratch (see Appendix~\ref{app:synthesize}).
    \ourmethod includes multiple iterations of fusion, where the synthesis of each iteration is based on the demonstration pool of the previous iteration.
    Each iteration consists of \textit{demonstration sampling} (\S\ref{subsec:sample}) and \textit{demonstration fusing} (\S\ref{subsec:fusion}) two steps, which sample and fuse the demonstrations of the demonstration pool separately.
    The fused demonstrations of each iteration are then added to the demonstration pool, preparing for the next iteration.

    After all iterations of fusion, we use the final demonstration pool for the text-to-SQL based on the in-context learning.
    We generate the SQL of each user question with LLMs directly following \citet{chang-fosler-lussier-2023-selective} since this is not the main topic of this paper.

\subsection{Demonstration Sampling}
    \label{subsec:sample}

    This step is designed to sample the demonstrations to be fused, which consists of:
    \textit{(i) Clustering} the demonstrations into multiple clusters;
    \textit{(ii) Sampling} demonstrations from clusters to be fused.

    \subsubsection{Clustering}
        \label{subsubsec:cluster}
        Before the fusion to get new demonstrations, it is required that the demonstrations sampled for fusing are not similar to ensure that the fused demonstration is not similar to the sampled demonstrations, thereby enhancing the diversity.
        The previous work~\cite{zhang2023automatic} has shown that similar demonstrations are in the same cluster after encoding and then clustering.
        That is because the encoded vectors can reflect the semantics of the demonstrations, where the closer the vector distance, the more similar the semantics.

        Inspired by this, we empirically employ an encoder model to encode the question of all demonstrations in the pool into vectors, and then use K-means to cluster encoded results into multiple clusters.
        Compared with not using the cluster, \ourmethod can ensure that the corresponding encoding vectors of the sampled demonstrations from different clusters are far away, leading to the demonstration used for fusion is not similar, enhancing diversity.

    \subsubsection{Sampling}
        \label{subsubsec:synthesis_demonstration}
        After obtaining different clusters of the demonstration pool, we then sample demonstrations from different clusters for fusing.
        Considering that even in a single cluster, there also exist differences between the demonstrations, since the encoded vector can not accurately reflect the complete information of the demonstration \cite{morris-etal-2023-loss_embedding}.
        To enhance diversity, during the demonstration sampling, we randomly choose several distinct clusters, and then randomly sample demonstrations from each cluster separately, making the fused demonstration reflect the difference between different demonstrations.

\subsection{Demonstration Fusing}
    \label{subsec:fusion}
    We employ LLM to fuse demonstrations as the discussion in Appendix~\ref{app:synthesize}, where we add the sampled demonstrations to guide the synthesis of the new demonstration as in-context learning, comparing with the randomly sampled database.
    Adding the sampled demonstrations comes up because LLMs imitate the demonstrations to generate results with the few-shot, whereas we let the LLM imitate both sampled demonstrations at the same time to get the fused demonstration.
    Thus, the fused demonstrations can reflect the attributes of and be different from all sampled demonstrations, thereby enhancing the diversity of the demonstration pool.

    \section{Experiments}
        \subsection{Experiment Setup}
    \label{subsec:experiment_setup}

    \paragraph{Dataset}
        We evaluate \ourmethod on two text-to-SQL datasets: Spider~\cite{yu-etal-2018-spider} and KaggleDBQA~\cite{lee-etal-2021-kaggledbqa}.
        Spider, a multi-domain text-to-SQL dataset, is one of the most widely used datasets currently.
        KaggleDBQA\footnote{We call KaggleDBQA as Kaggle for simplicity.} is smaller in scale but involves more complex database and SQL structures, presenting higher hardness.

    \paragraph{Metric}
        Following previous works~\cite{yu-etal-2018-spider,pourreza2023dinsql,li2023llm}, we employ execution match (EX) as our evaluation metric.
        EX measures the accuracy by comparing the execution results of the generated SQL on the database.
        There are two ways to evaluate EX: 
        \textit{(i)} directly using the predicted SQL conditional value (w. value);
        \textit{(ii)} replacing the conditional value with that in the correct SQL (w/o. value).

    \paragraph{Model}
        In our experiments, we use SGPT-125m~\cite{muennighoff2022sgpt} to encode demonstrations for clustering and use CodeLlama~\cite{rozière2023codellama} and GPT3.5\footnote{\href{https://platform.openai.com/docs/models/gpt-3-5}{Document} for GPT3.5.} to synthesize demonstrations and convert user questions into SQLs.
        We apply \ourmethod to the Vanilla method, ACT-SQL~\cite{zhang2023actsql} and ODIS~\cite{chang-fosler-lussier-2023-selective}, where the detail of these models and methods can be seen in Appendix~\ref{app:baseline}.

    \paragraph{Implementation Details}
        We study \ourmethod on two types of synthesis: from scratch (\textbf{\textit{w/o. Human}}) and based on human labeling (\textbf{\textit{w. Human}}).
        We synthesize $8$ SQLs for each given database, set the generation temperature to $0.3$, and synthesize in turns of $3$ (\textit{w/o. Human}) and $1$ (\textit{w. Human}) based on the analysis in \S~\ref{subsec:analysis_experiment}.
        About KaggleDBQA, we synthesize the demonstrations with both Spider and KaggleDBQA databases following the previous work \cite{chang-fosler-lussier-2023-selective}.
        The size of demonstration pools of different settings is shown in Appendix~\ref{app:synthesis_number}.
        We employ the 5-shot for text-to-SQL selected with BM-25 similarity, where the prompts for text-to-SQL are shown in Appendix~\ref{app:prompts}.

\subsection{Main Result}
    \begin{table*}[ht]
        \centering
        \small
        \begin{tabular}{lll|cc|cc|cc|cc|cc}
    \toprule
    \multirow{3}{*}{\textbf{Dataset}} & \multirow{3}{*}{\textbf{Method}} & \multirow{3}{*}{\textbf{Label}} & \multicolumn{6}{c}{\textbf{CodeLlama}} & \multicolumn{2}{c|}{\textbf{GPT3.5}} & \multicolumn{2}{c}{\bm{$\Delta$}} \\
     & & & \multicolumn{2}{c}{\textbf{7b}} & \multicolumn{2}{c}{\textbf{13b}} & \multicolumn{2}{c}{\textbf{34b}} & \multicolumn{2}{c|}{\textbf{-}} & \multicolumn{2}{c}{\textbf{-}} \\
     & & & w. & w/o. & w. & w/o. & w. & w/o. & w. & w/o. & w. & w/o. \\
    \midrule
    \multirow{9}{*}{Spider} & \multirow{5}{*}{Vanilla} & \textit{w/o. Human} & $48.5$ & $59.8$ & $54.9$ & $67.6$ & $56.9$ & $72.2$ & $57.9$ & $74.9$ & \multirow{2}{*}{$+3.4$} & \multirow{2}{*}{$+3.4$} \\
     & & \quad + \ourmethod & \textgreen{$54.4$} & \textgreen{$66.4$} & \textgreen{$58.8$} & \textgreen{$70.9$} & \textgreen{$59.7$} & \textgreen{$75.1$} & \textgreen{$58.7$} & \textgreen{$75.8$} & \\
    \cmidrule{3-13}
     & & \textit{w. Human} & $55.3$ & $67.5$ & $58.8$ & $72.1$ & $61.6$ & $76.7$ & $61.6$ & $80.3$ & \multirow{2}{*}{$+1.6$} & \multirow{2}{*}{$+1.4$} \\
     & & \quad + \ourmethod & \textgreen{$56.8$} & \textgreen{$69.0$} & \textgreen{$60.4$} & \textgreen{$74.2$} & \textgreen{$63.2$} & \textgreen{$78.4$} & \textgreen{$63.2$} & \textgreen{$80.7$} & \\
    \cmidrule{2-13}
     & \multirow{2}{*}{ACT-SQL$^{\dag}$} & \textit{w. Human} & \underline{$62.1$} & $63.2$ & $67.5$ & $69.1$ & $71.0$ & $72.8$ & $75.8$ & $77.6$ & \multirow{2}{*}{$+0.7$} & \multirow{2}{*}{$+0.9$} \\
     & & \quad + \ourmethod & \textred{$60.3$} & \textred{$61.7$} & \underline{\textgreen{$68.4$}} & \textgreen{$69.8$} & \underline{\textgreen{$74.6$}} & \textgreen{$76.7$} & \underline{\textgreen{$76.0$}} & \textgreen{$78.0$} & \\
    \cmidrule{2-13}
     & \multirow{2}{*}{ODIS$^{\dag}$} & \textit{w. Human} & $58.2$ & \underline{$71.8$} & $61.9$ & $76.6$ & $64.3$ & $80.9$ & $63.9$ & $81.1$ & \multirow{2}{*}{$+0.8$} & \multirow{2}{*}{$+0.9$} \\
     & & \quad + \ourmethod & \textred{$58.0$} & \textred{$71.0$} & \textgreen{$62.9$} & \underline{\textgreen{$78.0$}} & \textgreen{$65.6$} & \underline{\textgreen{$82.1$}} & \textgreen{$64.8$} & \underline{\textgreen{$83.0$}} & \\
    \midrule
    \multirow{9}{*}{Kaggle} & \multirow{5}{*}{Vanilla} & \textit{w/o. Human} & $9.9$ & $18.0$ & $13.2$ & $23.5$ & $13.2$ & $23.2$ & $14.0$ & $25.4$ & \multirow{2}{*}{$+6.1$} & \multirow{2}{*}{$+7.2$} \\
     & & \quad + \ourmethod & \textgreen{$22.8$} & \textgreen{$32.0$} & \textgreen{$19.1$} & \textgreen{$29.0$} & \textgreen{$18.0$} & \textgreen{$30.1$} & \textgreen{$14.7$} & \textgreen{$27.6$} & \\
    \cmidrule{3-13}
     & & \textit{w. Human} & $27.9$ & $39.7$ & $32.4$ & $44.1$ & $26.5$ & $38.6$ & $26.5$ & $40.4$ & \multirow{2}{*}{$+5.4$} & \multirow{2}{*}{$+4.2$} \\
     & & \quad + \ourmethod & \textgreen{$35.3$} & \underline{\textgreen{$47.1$}} & \textgreen{$34.6$} & \textgreen{$46.0$} & \textgreen{$32.4$} & \textgreen{$45.6$} & \textgreen{$32.4$} & \textgreen{$40.8$} & \\
    \cmidrule{2-13}
     & \multirow{2}{*}{ACT-SQL$^{\dag}$} & \textit{w. Human} & $27.6$ & $30.5$ & $30.5$ & $33.8$ & $33.8$ & $38.2$ & $29.4$ & $31.6$ & \multirow{2}{*}{$+0.4$} & \multirow{2}{*}{$+0.5$} \\
     & & \quad + \ourmethod & $27.6$ & \textgreen{$30.9$} & $30.5$ & $33.8$ & $33.8$ & \textgreen{$38.6$} & \textgreen{$30.9$} & \textgreen{$32.7$} & \\
    \cmidrule{2-13}
     & \multirow{2}{*}{ODIS$^{\dag}$} & \textit{w. Human} & $33.8$ & $43.4$ & $34.6$ & $47.1$ & $31.6$ & $46.3$ & $34.6$ & $48.9$ & \multirow{2}{*}{$+2.3$} & \multirow{2}{*}{$+3.0$} \\
     & & \quad + \ourmethod & \underline{\textgreen{$35.7$}} & \textgreen{$47.1$} & \underline{\textgreen{$36.0$}} & \underline{\textgreen{$48.5$}} & \underline{\textgreen{$35.3$}} & \underline{\textgreen{$50.4$}} & \underline{\textgreen{$36.8$}} & \underline{\textgreen{$51.5$}} & \\
    \bottomrule
\end{tabular}

        \caption{
            The main experimental results on the Spider and KaggleDBQA dev sets.
            About the label setting, \textit{w/o. Human} denotes synthesis from scratch using zero-shot and \textit{w. Human} denotes synthesis based on human labeling with few-shot.
            About the metric, w. denotes with values and w/o. denotes without values.
            $^{\dag}$ denotes the reproduced results since the performance differences brought by the API version of GPT3.5.
            The improved results led by \ourmethod are marked \textgreen{green}, the degradation is marked in \textred{red}, and unchanged results are marked in black.
            The best results of different models and datasets are annotated in \underline{underline}.
            \bm{$\Delta$} denotes the average improvement of different prompt methods leading by \ourmethod.
            We only adapt \textit{w/o. Human} to the Vanilla method since ACT-SQL and ODIS cannot be adapted to the zero-shot inference without labeling data.
        }
        \vspace{-1em}
        \label{tab:main_experiment}
    \end{table*}

    The text-to-SQL performance is shown in Table~\ref{tab:main_experiment}, where \ourmethod brings $3.2\%$ and $5.0\%$ performance improvement on average with and without human-labeling across different settings, showing the effectiveness of our method.
    We further discuss the performance under different SQL hardness in Appendix~\ref{app:sql_hardness}.
    From Table~\ref{tab:main_experiment}, we can also see that:

    \paragraph{Model Scale}
        Our method brings significant performance improvements on models of different scales.
        However, our method brings performance degradation with CodeLlama-7b, because of the low quality of the synthesized demonstrations due to the relatively poor performance of the 7b model, while ACT-SQL and ODIS are more sensitive to the demonstration quality since they employ the demonstrations to guide the intermediate generation rather than only for the few-shot.
        However, on KaggleDBQA, the performance does not increase as the model scale increases, because the demonstration pool used is synthesized or labeled with Spider databases (as described in \S~\ref{subsec:experiment_setup}), which could mislead the generation for the KaggleDBQA.

    \paragraph{Method}
        Our method continues to improve performance based on all experiment methods under most settings, even improving performance based on ODIS and ACT-SQL such two well-performed baselines, proving the generalization and effectiveness of \ourmethod.
        Compared to the Vanilla method, our method shows relatively minor improvements with ACT-SQL and ODIS.
        This is because ACT-SQL and ODIS are more effective in helping the model understand the reasoning process within demonstrations, rather than merely imitating.
        This reduces the dependency on the similarity between demonstrations and user questions, making performance improvements less sensitive to the diversity of the demonstration pool compared to Vanilla.

    \paragraph{Dataset}
        \ourmethod brings significant performance improvements on all experimental datasets and even achieves results close to \textit{w. Human} on Spider under the \textit{w/o. Human} setting, demonstrating the effectiveness of our method under different domains.
        Besides, our method significantly improves KaggleDBQA more than Spider, showing that the demonstrations synthesized by \ourmethod are more effective for complex text-to-SQL questions.

\subsection{Ablation Studies}
    \begin{table*}[t]
        \centering
        \small
        \begin{tabular}{l|ccc|ccc}
    \toprule
    \multirow{2}{*}{\textbf{Label}} & \multicolumn{3}{c}{\textbf{Spider}} & \multicolumn{3}{c}{\textbf{KaggleDBQA}} \\
     & \textbf{7b} & \textbf{13b} & \textbf{34b} & \textbf{7b} & \textbf{13b} & \textbf{34b} \\
    \midrule
    \ourmethod \textit{w/o. Human} & $66.4$ & $70.9$ & $75.1$ & $32.0$ & $29.0$ & $30.1$ \\
     \quad - Iteration & $66.2(-0.2)$ & $69.9(-1.0)$ & $73.9(-1.2)$ & $30.1(-1.9)$ & $28.8(-0.2)$ & $28.7(-1.4)$ \\
     \quad - Cluster & $65.3(-1.1)$ & $69.9(-1.0)$ & $74.6(-0.5)$ & $26.5(-5.5)$ & $26.5(-2.5)$ & $30.0(-0.1)$ \\
    \midrule
    \ourmethod \textit{w. Human} & $69.0$ & $74.2$ & $78.4$ & $47.1$ & $46.0$ & $45.6$ \\
     \quad - Iteration & $67.6(-1.4)$ & $71.9(-2.3)$ & $76.6(-1.8)$ & $38.6(-8.5)$ & $44.1(-1.9)$ & $38.6(-7.0)$ \\
     \quad - Cluster & $67.7(-1.3)$ & $70.5(-3.7)$ & $75.4(-3.0)$ & $41.2(-5.9)$ & $40.4(-5.6)$ & $35.7(-9.9)$ \\
    \bottomrule
\end{tabular}

        \caption{
            EX without values on CodeLlama ablating: \textit{(i) Iteration}: synthesizing the same demonstration number of \ourmethod in one single turn; \textit{(ii) Cluster}: randomly sampling demonstration to be fused without clustering.
        }
        \vspace{-1em}
        \label{tab:ablation}
    \end{table*}

    To verify the effectiveness of the iteration and the cluster designed by \ourmethod, we perform ablation experiments on each part separately.
    The experimental results are shown in Table~\ref{tab:ablation}.
    Based on such results, we discuss the impact of different parts on the performance of our method.

    \subsubsection{Ablation of Iteration}
        To demonstrate that iterations work by improving the quality rather than quantity of the demonstrations, we conduct experiments that generate the same number of data as our method without iterations.
        From Table~\ref{tab:ablation}, we can see that:
        \textit{(i)} There is a significant performance degradation after removing iteration, proving that \ourmethod enhances the performance by improving the demonstration quality rather than quantity;
        \textit{(ii)} For larger-scale models, iteration has a more significant impact on performance, indicating that larger-scale models can more effectively synthesize diverse demonstrations through multiple iterations;
        \textit{(iii)} Compared with \textit{w/o. Human}, \ourmethod under the \textit{w. Human} setting has a more obvious decrease after removing iteration, because the quality of the synthesis without labeling data is lower than the labeling data, mixing which leads to a quality degradation compared with the original labeling data.

    \subsubsection{Ablation of Cluster}
        To demonstrate the effectiveness of the cluster, we perform ablation experiments on it.
        We compare our method with randomly selecting demonstrations during the demonstration sampling.
        From Table~\ref{tab:ablation}, we can find:
        \textit{(i)} synthesis without clustering brings performance degradation in all settings, proving the effectiveness of the cluster;
        \textit{(ii)} The performance degradation of KaggleDBQA is more obvious compared to Spider, indicating that the more complex text-to-SQL questions are more sensitive to the demonstration diversity.

\subsection{Analysis}
    \label{subsec:analysis_experiment}

    In this part, we discuss the impact of different parameters on the model performance.
    The analysis experimental settings are shown in Appendix~\ref{app:analysis_setting}.

    \paragraph{Can Diversity Measurement Reflect the Diversity of the Demonstration Pool?}
        \begin{figure}[t]
            \centering
            \begin{tikzpicture}
    \small
    \begin{axis}[
        xlabel={\ourmetric},
        ylabel={EX w/o. values},
        ymin=71, ymax=76,
        xmin=0.05, xmax=0.10,
        xtick={0.05, 0.06, 0.07, 0.08, 0.09, 0.10},
        legend style={at={(0.5,-0.2)},anchor=north,legend columns=2},
        ymajorgrids=true,
        grid style=dashed,
        width=0.8\linewidth,
        every axis legend/.append style={nodes={anchor=west}},
        tick scale binop=\times,
        scaled ticks=false,
        xticklabel style={
            /pgf/number format/fixed,
            /pgf/number format/precision=3
        },
        yticklabel style={
            /pgf/number format/fixed,
            /pgf/number format/precision=1
        }
    ]

    \addplot[only marks, color=softblue, mark size=2pt, mark=*] coordinates {
        (0.058, 71.9)
        (0.065, 72.8)
        (0.066, 73.3)
        (0.069, 72.8)
        (0.069, 72.2)
        (0.069, 73.3)
        (0.069, 74.1)
        (0.071, 74.1)
        (0.072, 73.3)
        (0.072, 73.3)
        (0.078, 72.6)
        (0.081, 73.3)
        (0.083, 73.3)
        (0.084, 73.3)
        (0.084, 74.5)
        (0.086, 73.3)
        (0.090, 73.4)
        (0.093, 74.1)
        (0.093, 72.9)
        (0.096, 75.2)
    };

    \addplot [domain=0.05:0.10, samples=100, color=gray, thick] {70.401 + 38.101*x};

    \addplot [name path=upper, domain=0.05:0.10, samples=100, color=gray, dashed] {70.401 + 38.101*x + 1.141};
    \addplot [name path=lower, domain=0.05:0.10, samples=100, color=gray, dashed] {70.401 + 38.101*x - 1.141};

    \addplot [gray!10] fill between[of=upper and lower];

    \end{axis}
\end{tikzpicture}
            \vspace{-1em}
            \caption{
                EX of $20$ different demonstration pools with different \ourmetric on the Spider dev set.
                Different points denote different pools containing $100$ demonstrations randomly sampled from the Spider train set.
            }
            \vspace{-1em}
            \label{fig:diversity_label}
        \end{figure}
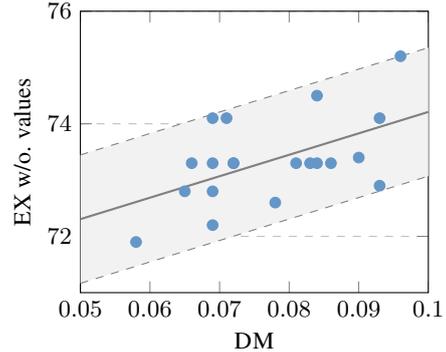
        
        To prove that the metric \ourmetric we proposed can reflect the diversity of the demonstration pool, we randomly sample $20$ demonstration pools, where each pool has $100$ demonstrations from the Spider train set with different diversities.
        Then we use the Vanilla method to evaluate the performance of each pool on the Spider dev set.
        The experiment results are shown in Figure~\ref{fig:diversity_label}, from which we can see that:
        \textit{(i)} With the same demonstration pool size, as \ourmetric enhances, the overall performance of the model is on the rise, indicating that the higher DM, the higher quality of the demonstration pool, denoting higher diversity;
        \textit{(ii)} Most of the results are concentrated around $73.3$, because such randomly sampled pools could not contain any demonstrations similar to the user questions, resulting in consistent performance.

    \paragraph{Does the Diversity and Performance of Synthesized Data Continue to Rise with the Iteration Turn Increasing?}
        \label{subsubsec:turn_number}

        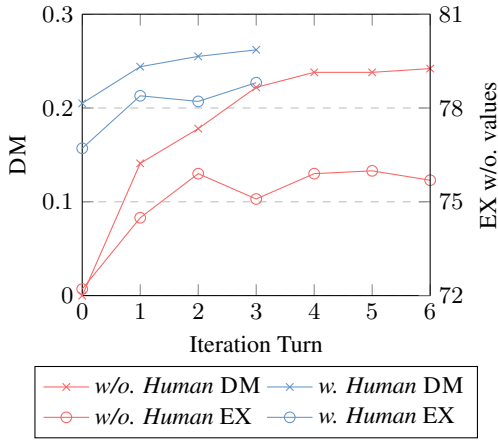
\begin{figure}[t]
            \centering
            \begin{tikzpicture}
    \small
    \begin{axis}[
        axis y line*=left,
        xlabel={Iteration Turn},
        xmin=0, xmax=6,
        ymin=0, ymax=0.3,
        ylabel={\ourmetric},
        ylabel near ticks,
        tick label style={font=\small},
        grid style=dashed,
        ymajorgrids=true,
        small,
        width=0.8\linewidth,
        legend style={at={(0.5,-0.25)},anchor=north, legend cell align={left}},
        legend columns=2
    ]
        \addplot[softred, mark=x] coordinates {
            (0, 0) (1, 0.141) (2, 0.178) (3, 0.222) (4, 0.238) (5, 0.238) (6, 0.242)
        };
        \addplot[softblue, mark=x] coordinates {
            (0, 0.205) (1, 0.244) (2, 0.255) (3, 0.262)
        };
        \addplot[softred, mark=o] coordinates {
            (0, 0.007) (1, 0.083) (2, 0.130) (3, 0.103) (4, 0.130) (5, 0.133) (6, 0.123)
        };
        \addplot[softblue, mark=o] coordinates {
            (0, 0.157) (1, 0.213) (2, 0.207) (3, 0.227)
        };
        \legend{\textit{w/o. Human} \ourmetric, \textit{w. Human} \ourmetric, \textit{w/o. Human} EX, \textit{w. Human} EX}
    \end{axis}
    
    \begin{axis}[
        axis y line*=right,
        axis x line=none,
        xmin=0, xmax=6,
        ymin=72, ymax=81,
        ylabel={EX w/o. values},
        ylabel near ticks,
        yticklabel style={font=\small},
        grid style=dashed,
        width=0.8\linewidth,
        ytick={72, 75, 78, 81}
    ]
    \end{axis}
\end{tikzpicture}
            \caption{
                \ourmetric and EX without values on the Spider dev set of CodeLlama-34b across different iterations with \ourmethod.
                Turn $0$ denotes the origin demonstration pool without \ourmethod.
                The sizes of the demonstration pools can be seen in Appendix~\ref{app:synthesis_number}.
            }
            \vspace{-1em}
            \label{fig:diversity_measurement}
        \end{figure}

        To analyze the effectiveness of the iteration, we adapt experiments with different iterative turns, which are summarized in Figure~\ref{fig:diversity_measurement}.
        From the table, we can see that: 
        \textit{(i)} When the turn is $\leq3$ (\textit{w/o. Human}) or $\leq1$ (\textit{w. Human}), as the turn increases, DM and the performance of our method improves steadily, indicating that multiple iterations can enhance the diversity, thereby enhancing performance;
        \textit{(ii)} When the turn is $>3$ (\textit{w/o. Human}) or $>1$ (\textit{w. Human}), with the number of turns increasing, diversity and performance improvement brought by \ourmethod becomes less and less, indicating the diversity can not be infinitely enhanced.
        Based on the above discussion, we use $3$ and $1$ as the synthesized turns.

    \paragraph{How Does the Synthesized Scale Effect the Performance}
        \begin{figure}[t]
            \centering
            \begin{tikzpicture}
    \small
    \begin{axis}[
        axis y line*=left,
        ylabel={Spider EX},
        xlabel={Synthesized Scale},
        xmin=0, xmax=4,
        ymin=71, ymax=79,
        xtick={0,1,2,3,4},
        tick label style={font=\small},
        xticklabels={0,10,100,1000,ALL},
        ytick={72, 74, 76, 78},
        legend style={at={(0.5,-0.25)},anchor=north,legend columns=2},
        ymajorgrids=true,
        grid style=dashed,
        width=0.8\linewidth,
        every axis legend/.append style={nodes={anchor=west}}
    ]
        \addplot[color=softred, mark=o] coordinates {(0,72.2) (1,73.3) (2,73.4) (3,74.0) (4,75.1)};
        \addlegendentry{Spider + \textit{w/o. Human}}

        \addplot[color=softblue, mark=o] coordinates {(0,71.1) (1,71.4) (2,71.5) (3,71.9) (4,73.4)};
        \addlegendentry{Kaggle + \textit{w/o. Human}}
        
        \addplot[color=softred, mark=x] coordinates {(0,76.7) (1,76.7) (2,77.1) (3,77.3) (4,78.4)};
        \addlegendentry{Spider + \textit{w. Human}}

        \addplot[color=softblue, mark=x] coordinates {(0,76.2) (1,76.7) (2,76.2) (3,77.0) (4,78.5)};
        \addlegendentry{Kaggle + \textit{w. Human}}
    \end{axis}

    \begin{axis}[
        axis y line*=right,
        axis x line=none,
        ylabel={KaggleDBQA EX},
        ymin=23, ymax=47,
        xmin=0, xmax=4,
        ylabel near ticks,
        ytick={26, 32, 38, 44},
        tick label style={font=\small},
        grid style=dashed,
        ymajorgrids=true,
        small,
        width=0.8\linewidth
    ]
    \end{axis}
\end{tikzpicture}
            \caption{
                The EX without values of CodeLlama-34b with different synthesized scales.
                The X-axis denotes the number of demonstrations randomly sampled from the synthesized data, where ALL denotes $1947$ and $10653$ demonstrations under the \textit{w/o. Human} and \textit{w. Human} respectively.
                The Y-axis on the left and right are the results of Spider and KaggleDBQA respectively.
            }
            \vspace{-1em}
            \label{fig:synthesis_scale}
        \end{figure}
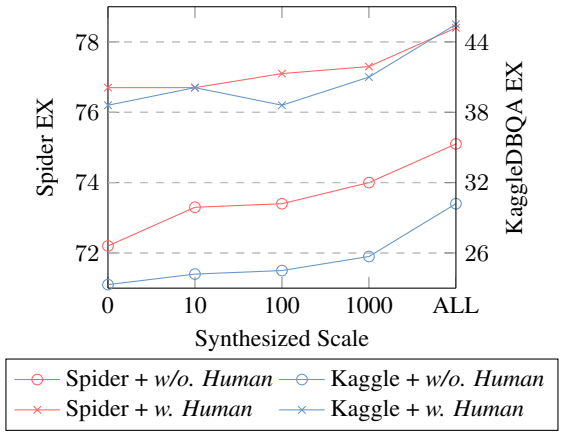

        To verify the impact of different synthesized scales on performance, especially the performance under the small synthesized scale, we adapt experiments on synthesizing different demonstration numbers.
        The experiment results are shown in Figure~\ref{fig:synthesis_scale}, from which we can see that:
        \textit{(i)} With the small synthesized scale ($\leq100$), \ourmethod can also improve the performance, proving the effectiveness under low synthesis overhead;
        \textit{(ii)} With the synthesized scale increasing, the performance is continuously enhancing, indicating that the synthesized scale has a significant impact on performance.

    \paragraph{How Does the Initial Labeling Scale Effect Our Synthesized Performance}
        \begin{figure}[t]
            \centering
            \begin{tikzpicture}
    \small
    \begin{axis}[
        axis y line*=left,
        xlabel={Initial Label Scale},
        ylabel={Spider EX},
        xmin=0, xmax=4,
        ymin=71, ymax=79,
        xtick={0,1,2,3,4},
        xticklabels={0,10,100,1000,7000},
        ytick={72, 74, 76, 78},
        tick label style={font=\small},
        legend style={at={(0.5,-0.25)},anchor=north,legend columns=2},
        ymajorgrids=true,
        grid style=dashed,
        width=0.8\linewidth,
        every axis legend/.append style={nodes={anchor=west}}
    ]
        \addplot[color=softred, mark=o] coordinates {(0,72.2) (1,73.2) (2,72.3) (3,74.7) (4,76.7)};
        \addlegendentry{Spider}
    
        \addplot[color=softblue, mark=o] coordinates {(0,71.1) (1,71.4) (2,72.8) (3,73.5) (4,76.2)};
        \addlegendentry{Kaggle}
        
        \addplot[color=softred, mark=x] coordinates {(0,74.5) (1,75.0) (2,76.3) (3,76.2) (4,78.4)};
        \addlegendentry{Spider + \ourmethod}
        
        \addplot[color=softblue, mark=x] coordinates {(0,72.3) (1,73.1) (2,73.2) (3,73.9) (4,78.5)};
        \addlegendentry{Kaggle + \ourmethod}
    \end{axis}

    \begin{axis}[
        axis y line*=right,
        axis x line=none,
        ylabel={KaggleDBQA EX},
        ymin=23, ymax=47,
        xmin=0, xmax=9,
        ylabel near ticks,
        ytick={26, 32, 38, 44},
        tick label style={font=\small},
        grid style=dashed,
        ymajorgrids=true,
        small,
        width=0.8\linewidth
    ]
    \end{axis}
\end{tikzpicture}
            \caption{
                The EX without values of CodeLlama-34b under different initial human labeling scales sampled from the Spider train set.
                The X-axis represents the number of labeled demonstrations used for synthesis.
                The Y-axis on the left and right represent the results of Spider and KaggleDBQA respectively.
            }
            \vspace{-0.5em}
            \label{fig:label_scale}
        \end{figure}
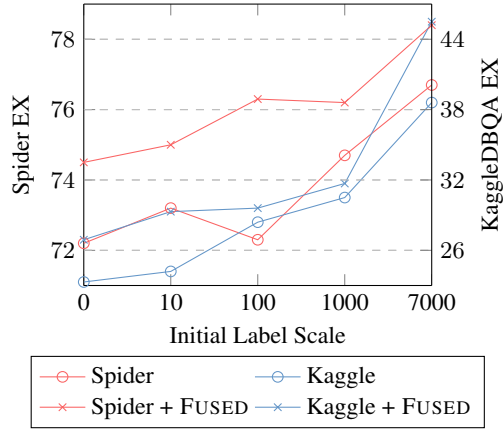
    
        Although the main experiments of Table~\ref{tab:main_experiment} demonstrate the effectiveness of our method on labeled data, the practical applications could lack labeled data with the same scale as the Spider training data.
        Therefore, to validate the effectiveness of \ourmethod across varying scales of labeling, we randomly sample and conduct experiments on initial labeling demonstrations of different numbers from Spider training data.

        The experiment results are shown in Figure~\ref{fig:label_scale}, from which we can see that:
        \textit{(i)} Under most settings, our method brings performance improvement, indicating its widespread effectiveness under different initial label scales;
        \textit{(ii)} With the increase of the initial label scale, the performance demonstrates a consistent increase, suggesting that expanding the labeling scale can reliably enhance performance.

    \paragraph{Can \ourmethod Effectively Help LLMs Migrate to the Domain without Labeling?}
        \begin{table}[t]
            \centering
            \small
            \begin{tabularx}{\linewidth}{l|CCC}
                \toprule
                \textbf{Database} & \textbf{7b} & \textbf{13b} & \textbf{34b} \\
                \midrule
                None & $18.0$ & $23.5$ & $23.2$ \\
                Kaggle & $29.0$ & $24.3$ & $27.6$ \\
                Kaggle + Spider & $32.0$ & $29.0$ & $30.1$ \\
                \bottomrule
            \end{tabularx}
            \caption{
                EX without values of \ourmethod using CodeLlama evaluated on the KaggleDBQA dev set with the data synthesized based on the databases of different datasets under the \textit{w/o. Human} setting.
                None denotes no synthesis data, Kaggle denotes synthesis only with the KaggleDBQA databases, and Kaggle + Spider denotes synthesis by mixing Spider databases.
            }
            \vspace{-1em}
            \label{tab:database_domain}
        \end{table}

        In this part, we evaluate that \ourmethod can improve the text-to-SQL performance across different domains without human labeling.
        The experimental results are shown in Table~\ref{tab:database_domain}.
        From the table, we can see that:
        \textit{(i)} Compared with not synthesizing demonstrations, \ourmethod can bring performance improvements when only using KaggleDBQA databases, proving the effectiveness of our method adapted to a new domain without labeling;
        \textit{(ii)} Compared to using only KaggleDBQA databases, the demonstrations obtained by mixing Spider databases can bring greater performance improvements, indicating that increasing the diversity of databases can also enhance the diversity of synthesized demonstrations.

\subsection{Case Study}
    \begin{figure}
        \centering
        \includegraphics[width=0.9\linewidth]{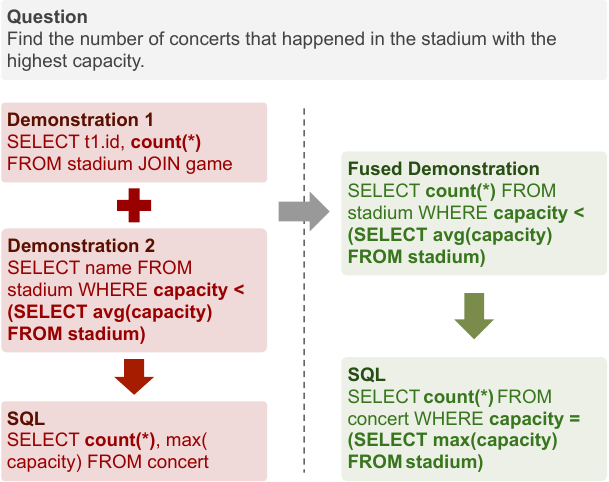}
        \caption{
            The case study of demonstrations by human-labeling (left) and \ourmethod (right) from Spider.
            The corresponding SQL keywords between demonstrations and the answer are annotated in \textbf{bold}.
        }
        \vspace{-1em}
        \label{fig:case_study}
    \end{figure}

    Although the above analysis proves the effectiveness of \ourmethod, how our method improves the performance of the text-to-SQL using in-context learning remains to be discovered.
    To analyze how our method improves the model performance more specifically, in this part, we conduct a case study.
    A comparison between results based on labeled data and the demonstrations obtained using \ourmethod is shown in Figure~\ref{fig:case_study}.
    From the figure, we can see that the results using only labeled data do not combine the SQL keywords of the two demonstrations well.
    The demonstration obtained with our method, on the other hand, has already combined the SQL keywords of the two demonstrations, which guides the model to successfully generate the correct SQL.

    \section{Related Works}
        \subsection{Text-to-SQL}
    Text-to-SQL is a vital task that generates SQL based on the user question and the provided databases.
    Recent research shows that text-to-SQL based on LLMs can approach or exceed the performance of fine-tuned models without fine-tuning, which greatly advances research on this task while reducing labeling overhead \cite{chang-fosler-lussier-2023-selective,zhang2023actsql,li2024using}.
    For example, DIN-SQL~\cite{pourreza2023dinsql} decomposes the text-to-SQL task into multiple sub-tasks.
    DAIL-SQL~\cite{gao2023dailsql} evaluates different prompt formats to find the best combination.
    MCS-SQL~\cite{lee2024mcssql} consistency the results generated with multiple prompts.

    However, existing LLM-based methods entirely rely on human-labeled demonstrations, demanding high labeling costs be adapted to a new domain.
    Therefore, we propose \ourmethod to synthesize text-to-SQL demonstrations based on LLMs using provided domain databases without human labeling, effectively reducing the labor cost.

\subsection{In-Context Learning}
    In-context learning is an effective method to enhance the reasoning ability of LLMs by providing several demonstrations to guide reasoning \cite{in-context-survey,wei2022ChainOfThought}.
    Some works propose to automatically select relevant demonstrations for each user question to improve the performance of LLMs \cite{zhang2023automatic,shum-etal-2023-automatic,qu2024deepicl}.
    Another kind of work enhances in-context learning by synthesizing relevant data by supervised fine-tuning \cite{wang-etal-2023-self-instruct,yang2024selfdistillation,sun2023SelfAlign}.

    However, existing methods only demonstrate that increasing the diversity of the demonstrations can enhance performance but do not discuss if the diversity of the existing labeling data is sufficient, and how to increase the diversity of the demonstrations \cite{su2023selective,levy-etal-2023-diverse}.
    Therefore, we present \ourmetric to show that the existing labeling data of the text-to-SQL is not diverse enough and propose \ourmethod to enhance the diversity.

    \section{Conclusion}
        In this paper, we improve the performance of the text-to-SQL task using in-context learning from the perspectives of measuring and enhancing the demonstration pool diversity.
        We first present \ourmetric to measure the diversity of the demonstration pool, based on which we present that the diversity of the existing labeling data can be further enhanced.
        Based on the above analysis, we present \ourmethod, which synthesizes demonstrations using LLMs, lowering the labeling cost.
        Experiments show that \ourmethod brings an average improvement of $3.2\%$ and $5.0\%$ with and without labeling data on Spider and KaggleDBQA, proving the effectiveness.

    \newpage
    \section*{Limitations}
        \ourmethod has two limitations, including:
        \textit{(i)} About the encoding of the demonstration sample step, directly splice the user question and the SQL could not fully reflect the attributes of them.
        In future work, we will try to encode the question and SQL according to the attributes separately;
        \textit{(ii)} For the synthesized demonstration pool, we only enhance the diversity, while ignoring the effect of the scale on the demonstration selection.
        Our future work will filter the synthesis, reducing the scale of synthesis under the premise of ensuring diversity.

    \section*{Ethics Statement}
        All datasets and models used in this paper are publicly available, and our usage follows their licenses and terms.

    \bibliography{custom}

    \clearpage
    \appendix
    \section{Comparison with Other Diversity Metrics}
    \label{app:other_metric_comparison}

    To better explore the progress of the diversity metric we proposed in \S\ref{sec:discussion}, we compare it with the past metric present by \citet{Nan2023EnhancingFT}:
    \textit{(i)} \citet{Nan2023EnhancingFT} mainly focus on selecting demonstrations, while the motivation of ours is to synthesize demonstrations; 
    \textit{(ii)} \citet{Nan2023EnhancingFT} does not give a numerical measure of diversity, while our method gives a numerical measure of diversity; 
    \textit{(iii)} \citet{Nan2023EnhancingFT} is based on clustering, and the granularity of judging diversity is relatively coarse, while our method is based on the entire demonstration pool, which can more accurately measure the diversity of demonstrations.

\section{How to Calculate Diversity Measurement}
    \label{app:measurement}

    \begin{equation}
        \begin{aligned}
            &\min_{u \in U} \max_{d_i \in D} \texttt{sim}(u, d_i)\\
            &\coloneqq \min_{u \in \texttt{Convex}(D)} \max_{d_i \in D} |u - d_i|^{-1}
        \end{aligned}
        \label{equ:optimization}
    \end{equation}

    As the discussion in \S\ref{sec:discussion}, given a demonstration pool, the calculation process of \ourmetric can be formalized as Equation~\ref{equ:optimization}, where $\texttt{sim}(u, d) = |u - d|^{-1}$ is the reciprocal of the Euclidean distance between the encoded vectors of $u$ and $d$, and $\texttt{Convex}(D)$ denotes the convex hull of the demonstrations.

    We use the Euclidean distance to represent $\texttt{sim}$ since the closer the distance between the embedding question and the embedding demonstration, the more similarity between the question and the demonstration.
    The user question $u$ should be in the area surrounded by $\texttt{Convex}(D)$ corresponds to the question-related domain, and the user questions are highly related to the domain and have a high probability of locating in the convex.

    We use SciPy~\cite{2020SciPy-NMeth} to solve Equation~\ref{equ:optimization}, and use SGPT-125m \cite{muennighoff2022sgpt} to encode demonstrations.
    We first generate the Voronoi diagram~\cite{voronoi} and compute the convex hull for the encoded demonstration points.
    For each point, we then calculate the maximum distance to any vertex in its corresponding Voronoi region confined within the convex hull and use the greatest of these maximum distances as the result.

\section{Text-to-SQL Prompts}
    \label{app:prompts}

    The prompts of the SQL generation and the question generation are shown in Table~\ref{tab:sql_generation_prompt} and Table~\ref{tab:question_generation_prompt}, where the formats of \texttt{\{database\}} and \texttt{\{demonstration\}} are same as \citet{chang-fosler-lussier-2023-selective}.

\section{Why \ourmethod can Enhance the Diversity Measurement}
    \label{app:provment_of_improvement}

    In this section, we explain why the demonstration sampling (\S\ref{sec:method}) in \ourmethod can enhance \ourmetric.
    To increase Equation~\ref{equ:diversity_measurement}, it is required to maximum $\min_{u \in U}\ \max_{d_i \in D} \texttt{sim}(u, d_i)$.
    Let $u^{*} = \argmin_{u \in U}\ \max_{d_i \in D} \texttt{sim}(u, d_i)$, then we aim to update $D$ to make $\max_{d_i \in D} \texttt{sim}(u^{*}, d_i)$ as large as possible.

    We define that the cluster corresponding to $d_i$ is $c_{d_i}$, and let $\texttt{sim}(u, c_i) = \max_{d \in c_i} \texttt{sim}(u, d)$.
    We denote $d_i$ that maximum $\texttt{sim}(u^{*}, d_i)$ as $d^{*}$.
    Then we have $\max_{d_i \in D} \texttt{sim}(u^{*}, d_i) = \texttt{sim}(u^{*}, c_{d^{\dag}}) = |u^{*} - c_{d^{*}}|^{-1} > (|u^{*} - c| + |c - c_{d^{*}}|)^{-1}$, where $c$ is any cluster.
    The above inequality holds because $c_{d^{\dag}}, c, u^{*}$ can be considered as the vertices of a triangle, and the sum of the lengths of two sides is greater than the length of the third side.

    According to the discussion in Appendix~\ref{app:measurement}, it is hard to precisely find $u^{*}$, so we maximize the right-hand side of the inequality as much as possible to increase $\texttt{sim}(u^{*}, c_{d^{*}})$.
    Therefore, as described in \S\ref{subsec:sample}, the demonstration sampling continuously combines demonstrations to generate new demonstrations between different clusters, thereby reducing the distance between different clusters.
    During the sampling, adding new results can also decrease the distance between $u^{*}$ and $c$, so the right-hand side of the inequality is continuously decreasing.
    In summary, \ourmethod can continuously increase $\max_{d_i \in D} \texttt{sim}(u^{*}, d_i)$, thus increasing \ourmetric of the results.

\section{Synthesize Text-to-SQL Demonstrations with LLMs}
    \label{app:synthesize}
    
    In this section, we discuss how to employ LLMs to obtain the initial demonstration pool with the given database, lowering the labeling cost. 
    The prompts we used are shown in Appendix~\ref{app:prompts}.

    \begin{table}[t]
        \centering
        \small
        \begin{tabular}{p{0.9\linewidth}}
            \toprule
            \textbf{\textit{\underline{SQL Synthesize}}} \\
            Synthesize one SQL query for the given database. \\
            \\
            \{database\} \\
            -- Synthesize a new single SQL for the above database imitating \{SQL1\} and \{SQL2\}. \\
            SELECT \\
            \bottomrule
        \end{tabular}
        \caption{The prompt for the SQL synthesis.}
        \label{tab:sql_generation_prompt}
    \end{table}

    \paragraph{SQL Synthesize}
        Following the previous work~\cite{chang-fosler-lussier-2023-selective}, we synthesize SQL based on the linearized schema of the given database with LLMs.
        During synthesis, we ask LLMs to generate multiple SQLs for each database to enhance the diversity of the results with the sampling generation.
        The prompt we used is shown in Table~\ref{tab:sql_generation_prompt}. 

    \begin{table}[t]
        \centering
        \small
        \begin{tabular}{p{0.9\linewidth}}
            \toprule
            \textbf{\textit{\underline{Question Synthesize}}} \\
            Using natural language, generate a question corresponding to the given SQL. \\
            Different examples are separated with `\textbackslash n\textbackslash n'. \\
            \\
            \{demonstration1\} \\
            \\
            ... \\
            \\
            \{demonstration5\} \\
            \\
            \{database\} \\
            -- Using natural language, generate a question corresponding to the given SQL: \{SQL\}.\\
            Question: \\
            \bottomrule
        \end{tabular}
        \caption{The prompt for the question synthesis.}
        \label{tab:question_generation_prompt}
    \end{table}

    \paragraph{Question Synthesize}
        We synthesize the corresponding questions of the generated SQL with the linearized schema of the.
        We first synthesize SQL instead of questions because LLMs could generate questions that are hard to answer using SQL \cite{cheng2023binding}, and it is harder to validate the semantic consistency between the SQL and the question for generating questions first.
        The prompt of this step is shown in Table~\ref{tab:question_generation_prompt}.

    \paragraph{Validate}
        Due to the limitation of the model performance, it is hard to guarantee that the semantics of all synthesized SQL-question pairs are completely consistent, resulting in a decrease in the quality of the synthesized demonstration.
        To improve the quality of the synthesized results, we verify the semantic consistency between the synthesized questions and SQL. 
        We generate SQL based on the question and then evaluate if the generated SQL is the same as the synthesized SQL, for which we use LLMs to reduce the cost of fine-tuning.
        The prompts for text-to-SQL follow \citet{chang-fosler-lussier-2023-selective}.

\section{Baselines}
    \label{app:baseline}

    \subsection{Baseline Models}
        \paragraph{CodeLlama}
            CodeLlama is a model based on Llama2~\cite{touvron2023llama2}, which is fine-tuned on a large amount of code data and can better solve code-related problems (including SQL).
    
        \paragraph{GPT3.5}
            GPT3.5 is an improved model based on GPT3~\cite{gpt3}, which further enhances performance through additional task-specific fine-tuning.
            We use Azure OpenAI API of \texttt{gpt-3.5-turbo} of GPT3.5 for our experiments~\footnote{\url{https://azure.microsoft.com/en-us/products/cognitive-services/openai-service}}.

    \subsection{Baseline Methods}
        \paragraph{Vanilla}
            Following the previous work \cite{chang-fosler-lussier-2023-selective}, we design the Vanilla method that directly employs the few-shot to generate the answer, where the demonstrations are selected by the BM-25 similarity between the user question and the demonstration questions.
    
        \paragraph{ACT-SQL}
            ACT-SQL~\cite{zhang2023actsql} is a method to construct the chain-of-thought rationales based on SQL automatically.
            This method synthesizes reasoning steps with table names, column names, and values used in the SQL.
    
        \paragraph{ODIS}
            ODIS~\cite{chang-fosler-lussier-2023-selective} is an automatic demonstration selection method designed for the text-to-SQL task.
            This method selects out-domain demonstrations from the labeled data and synthesizes in-domain demonstrations based on the databases related to the user question.

\section{Number of Synthesized Data}
    \label{app:synthesis_number}

    \begin{table*}[ht]
        \small
        \centering
        \begin{tabular}{llcccccc}
            \toprule
            \multirow{2}{*}{\textbf{Model}} & \multirow{2}{*}{\textbf{Label}} & \multicolumn{4}{c}{\textbf{Turn}} & \multirow{2}{*}{\textbf{Total}} \\
             & & \textbf{0} & \textbf{1} & \textbf{2} & \textbf{3} & \\
            \midrule
            \multirow{2}{*}{CodeLlama-7b} & \textit{w/o. Human} & $0$ & $584$ & $561$ & $608$ & $1753$ \\
             & \textit{w. Human} & $7000$ & $1937$ & $-$ & $-$ & $8937$ \\
            \midrule
            \multirow{2}{*}{CodeLlama-13b} & \textit{w/o. Human} & $0$ & $954$ & $741$ & $803$ & $2489$ \\
             & \textit{w. Human} & $7000$ & $3001$ & $-$ & $-$ & $10001$ \\
            \midrule
            \multirow{2}{*}{CodeLlama-34b} & \textit{w/o. Human} & $0$ & $457$ & $668$ & $822$ & $1947$ \\
             & \textit{w. Human} & $7000$ & $3653$ & $-$ & $-$ & $10653$ \\
            \midrule
            \multirow{2}{*}{\texttt{gpt-3.5-turbo}} & \textit{w/o. Human} & $0$ & $803$ & $643$ & $502$ & $1948$ \\
             & \textit{w. Human} & $7000$ & $387$ & $-$ & $-$ & $7387$ \\
            \bottomrule
        \end{tabular}
        \caption{Synthesized size under different settings.}
        \label{tab:pool_size}
    \end{table*}

    The synthesized demonstrations under different settings are shown in Table~\ref{tab:pool_size}.
    From the table, we can see that \texttt{gpt-3.5-turbo} has less data than that synthesized by CodeLlama, because the SQL synthesized by \texttt{gpt-3.5-turbo} is more complex, which makes it more difficult to pass the filter.

    \begin{table*}[ht]
        \small
        \centering
        \begin{tabular}{llccccccccc}
            \toprule
            \multirow{2}{*}{\textbf{Model}} & \multirow{2}{*}{\textbf{Label}} & \multicolumn{7}{c}{\textbf{Turn}} \\
             & & \textbf{0} & \textbf{1} & \textbf{2} & \textbf{3} & \textbf{4} & \textbf{5} & \textbf{6} \\
            \midrule
            \multirow{2}{*}{CodeLlama-34b} & \textit{w/o. Human} & $0$ & $457$ & $668$ & $822$ & $854$ & $981$ & $897$ \\
             & \textit{w. Human} & $7000$ & $3653$ & $4243$ & $4344$ & $-$ & $-$ & $-$ \\
            \bottomrule
        \end{tabular}
        \caption{More synthesized size of CodeLlama-34b for Figure~\ref{fig:diversity_measurement}.}
        \label{tab:pool_size_extend}
    \end{table*}

    To find the best turn number of synthesis, we synthesize more turns on CodeLlama-34b, and the size of synthetic data is shown in Table~\ref{tab:pool_size_extend}.

\section{\ourmethod Performance under Different SQL Hardness}
    \label{app:sql_hardness}

    \begin{table}[t]
        \centering
        \small
        \begin{tabularx}{\linewidth}{ll|CCCC}
            \toprule
            \textbf{Dataset} & \textbf{Label} & \textbf{Easy} & \textbf{Medium} & \textbf{Hard} & \textbf{Extra} \\
            \midrule
            \multirow{2}{*}{Spider} & \textit{w/o. Human} & \bm{$88.7$} & $80.7$ & $57.5$ & $40.4$ \\
             & \quad + \ourmethod & $87.5$ & \bm{$81.2$} & \bm{$63.8$} & \bm{$52.4$} \\
            \midrule
            \multirow{2}{*}{Kaggle} & \textit{w/o. Human} & $53.1$ & $30.3$ & $5.1$ & \textbf{$1.9$} \\
             & \quad + \ourmethod & \bm{$59.4$} & \bm{$32.9$} & \bm{$11.4$} & \bm{$1.9$} \\
            \bottomrule
        \end{tabularx}
        \caption{
            EX without values of CodeLlama-34b under different SQL hardness with and without \ourmethod.
            The best result of each setting is annotated in \textbf{bold}.
        }
        \label{tab:sql_hardness}
    \end{table}

    To analyze the effectiveness of \ourmethod on questions with different complexity, we evaluate our method on SQL categorized by different hardness.
    The category criteria follows \citet{yu-etal-2018-spider}.
    The experimental results are shown in Table~\ref{tab:sql_hardness}.

    From the table, we can see that:
    \textit{(i)} On most hardness, our method can bring significant performance improvements, which proves the effectiveness of \ourmethod;
    \textit{(ii)} On Spider, the more difficult SQL, the more significant the improvement, showing that synthesized demonstrations can more effectively guide complex SQL generation;
    \textit{(iii)} For the easy questions of Spider, our method brings a slight performance degradation because the model already performs well under the \textit{w/o. Human} setting for this hardness, and the additional demonstrations could mislead the model;
    \textit{(iv)} On the extra questions of KaggleDBQA, our method does not bring performance improvement, which could be because it is too hard to synthesize too complex demonstrations (harder than Spider extra questions), resulting in the selected demonstrations being unable to effectively guide the generation of the extra hardness.

\section{Settings of Analysis Experiments}
    \label{app:analysis_setting}

    We adapt analysis experiments under the setting of:
    \paragraph{CodeLlama-34b} CodeLlama is one of the most mainstream code generation models at present, which achieves near the performance of the closed-source model (as shown in Table~\ref{tab:main_experiment}) in the open-source model with less inference cost (no need to call API), of which CodeLlama-34b is the best performance in this series of models.
    \paragraph{Evaluating without values} Regarding the text-to-SQL task, current research mainly focuses on how to generate SQL with the correct structure, while paying less attention to extracting the condition values exactly, since this requires the memorizing ability rather than the semantic parsing ability.

\section{Synthesized Template}
    \begin{table}[t]
        \centering
        \small
        \resizebox{\linewidth}{!}{
            \begin{tabular}{p{\linewidth}}
                \toprule
                \textbf{Template (\%)} \\
                \midrule
                \texttt{SELECT * FROM * WHERE * <op> *} $(25.7)$ \\
                \texttt{SELECT * FROM * WHERE * <op> * AND * <op> *} $(13.9)$ \\
                \texttt{SELECT * FROM * JOIN * JOIN * WHERE * <op> *} $(5.2)$ \\
                \texttt{SELECT * FROM * JOIN * WHERE * <op> *} $(4.9)$ \\
                \texttt{SELECT * FROM * WHERE * IN (SELECT * FROM * WHERE * <op> *)} $(4.3)$ \\
                \bottomrule
            \end{tabular}
        }
        \caption{
            Top five SQL templates synthesized by \ourmethod using CodeLlama-34b.
            The numbers in the brackets denote the proportion of each template.
        }
        \label{tab:sql_template}
    \end{table}

    To guide future works in generating more diverse demonstrations, in this part, we analyze the proportion of demonstrations with different SQL templates synthesized by our method.
    We replace table names, column names, and values with \texttt{*} and operators with \texttt{<op>} as the templates corresponding to each SQL.
    Our method synthesizes $175$ different SQL templates, showing the diversity of the synthesized demonstrations.
    The five most frequent template types are shown in Table~\ref{tab:sql_template}.

    From the table, we can find:
    \textit{(i)} The current model is most inclined to generate \texttt{SELECT} and \texttt{WHERE}, which is nearly $40\%$, indicating that such types of SQL occur more frequently in the pre-training data of LLMs we use and, thereby, are more frequently used in real-world scenarios;
    \textit{(ii)} Existing models hardly generate complex SQL that contains nested SQL (less than $5\%$ of synthetic data), indicating that future methods should specifically pay attention to guide the model to generate results that contain two or more sub-SQLs or even more complex structures.

\end{document}